\documentclass[wcp]{jmlr}
\usepackage{longtable}% for long tables

\usepackage[load-configurations=version-1]{siunitx} % newer version
\usepackage{psfrag}

\newcommand{\expect}{\mathbf{E}}
\newcommand{\reals}{\mathbb{R}}

\newcommand{\comment}[1]{}
\jmlrvolume{vol}
\jmlryear{2012}
\jmlrworkshop{EWRL 2012}

\title[Generalized Optimality Equations]{Free Energy and the Generalized Optimality Equations for Sequential Decision Making}

\author{%
\Name{Pedro A. Ortega}
\Email{pedro.ortega@tuebingen.mpg.de}\\
\addr Max Planck Institute for Biological Cybernetics\\
Max Planck Institute for Intelligent Systems\\
ORAND S.A.\\
\AND
\Name{Daniel A. Braun} \Email{daniel.braun@tuebingen.mpg.de}\\
\addr Max Planck Institute for Biological Cybernetics\\
Max Planck Institute for Intelligent Systems}

\editor{Marc Deisenroth, Csaba Szepesvari, Jan Peters}

\begin{document}

\maketitle

%%%%%%%%%%%%%%%%%%%%%%%%%%%%%%%%%%%%%%%%%%%%%%%%%%%%%%%%%%%%%%%%%%%%%%
% Body of Article                                                    %
%%%%%%%%%%%%%%%%%%%%%%%%%%%%%%%%%%%%%%%%%%%%%%%%%%%%%%%%%%%%%%%%%%%%%%

\begin{abstract}
The free energy functional has recently been proposed as a variational principle for bounded rational decision-making, since it instantiates a natural trade-off between utility gains and information processing costs that can be axiomatically derived. Here we apply the free energy principle to general decision trees that include both adversarial and stochastic environments. We derive generalized sequential optimality equations that not only include the Bellman optimality equations as a limit case, but also lead to well-known decision-rules such as Expectimax, Minimax and Expectiminimax. We show how these decision-rules can be derived from a single free energy principle that assigns a resource parameter to each node in the decision tree. These resource parameters express a concrete computational cost that can be measured as the amount of samples that are needed from the distribution that belongs to each node. The free energy principle therefore provides the normative basis for generalized optimality equations that account for both adversarial and stochastic environments.
\end{abstract}


%%%%%%%%%%%%%%%%%%%%%%%%%%%%%%%%%%%%%%%%%%%%%%%%%%%%%%%%%%%%%%%%%%%%%%
% Body of Article                                                    %
%%%%%%%%%%%%%%%%%%%%%%%%%%%%%%%%%%%%%%%%%%%%%%%%%%%%%%%%%%%%%%%%%%%%%%

\section{Introduction}

Decision trees are a ubiquitous tool in decision theory and artificial intelligence research to represent a wide range of decision-making problems that include the classic reinforcement learning paradigm as well as competitive games \citep{Osborne1999,RussellNorvig2010}. Depending on the kind of system one is interacting with, there are different decision rules
one has to apply---the most famous ones being \emph{Expectimax}, \emph{Minimax} and \emph{Expectiminimax}---see~Figure~\ref{fig:gametrees}. When an agent interacts with a stochastic system, the agent chooses its decisions based on \emph{Expectimax}. Essentially, Expectimax is the dynamic programming algorithm that solves the Bellman optimality equations, thereby recursively maximizing expected future reward in a sequential decision problem \citep{Bellman1957}.

In two-player zero-sum games where strictly competitive players make alternate moves, an agent should use the \emph{Minimax} strategy. The motivation underlying minimax decisions is that the agent wants to optimize the worst-case gain as a means of protecting itself against the potentially harmful decisions made by the adversary. Finally, there are games that mix the two previous interaction types. For instance, in Backgammon, the course of the game depends on the skill of the players and chance elements. In these cases, the agent bases its decisions on the \emph{Expectiminimax} rule \citep{Michie1966}.

\begin{figure}[htbp]
\floatconts
  {fig:gametrees}
  {\caption{Illustration of Expectimax, Minimax and Expectiminimax in decision trees representing three different interaction scenarios. The internal nodes can be of three possible types: maximum ($\vartriangle$), minimum ($\triangledown$) and expectation ($\circ$). The optimal decision is calculated recursively using dynamic programming.}}
  {\footnotesize
  \psfrag{l1}[c]{Expectimax}
  \psfrag{l2}[c]{Minimax}
  \psfrag{l3}[c]{Expectiminimax}
  \psfrag{a1}[c]{$\max$}
  \psfrag{a2}[c]{$\mathbf{E}$}
  \psfrag{a3}[c]{$\max$}
  \psfrag{a4}[c]{$\mathbf{E}$}
  \psfrag{a5}[c]{$\max$}
  \psfrag{a6}[c]{$\min$}
  \psfrag{a7}[c]{$\max$}
  \psfrag{a8}[c]{$\min$}
  \psfrag{a9}[c]{$\mathbf{E}$}
  \psfrag{ab}[c]{$\max$}
  \psfrag{ac}[c]{$\mathbf{E}$}
  \psfrag{ad}[c]{$\min$}
  \includegraphics[]{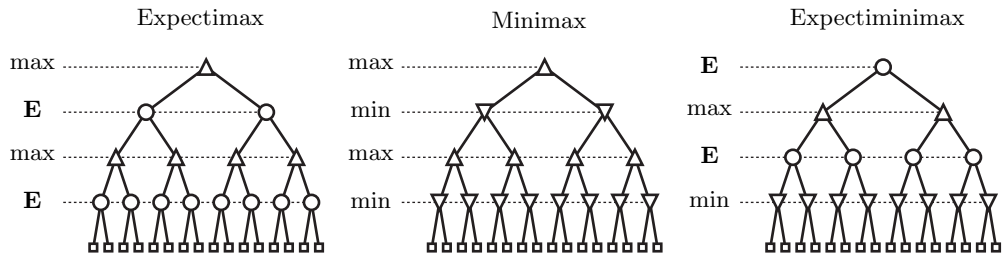}}
\end{figure}

What is common to all of these decision-making schemes is that they presuppose a fully rational decision-maker that is able to compute all of the required operations with absolute precision. In contrast, a bounded rational decision-maker trades off expected utility gains against the
cost of the required computations  \citep{Simon1984}. Recently, the free energy has been suggested as a normative variational principle for such bounded rational decision-making that takes the computational effort into account \citep{OrtegaBraun2011b,BraunOrtega2011,Ortega2011}. This builds on previous work on efficient computation of optimal actions that trades off the benefits obtained from maximizing the utility function against the cost of changing the uncontrolled dynamics given by the environment \citep{Kappen2005,Todorov2006,Todorov2009,Kappen2012}. The aim of this paper is to extend these results to generalized decision trees such that
Expectimax, Minimax, Expectiminimax, and bounded rational acting can all be derived from a single optimization principle. Moreover, this framework leads to a natural measure of computational costs spent at each node of the decision tree. All the proofs are given in the appendix.

\section{Free Energy}

\subsection{Equilibrium Distribution}
In~\citet{OrtegaBraun2011b} and in~\citet{Ortega2011} it was shown that a bounded rational decision-making problem can be formalized based on the \emph{negative free energy difference} between two information processing states represented by two probability distributions~$P$ and~$Q$. The decision process then transforms the initial choice probability~$Q$ into a final choice probability~$P$ by taking into account the utility gains (or losses) and the transformation costs. This transformation process can be formalized as
\begin{equation}\label{eq:equilibrium}
    P(x) = \frac{1}{Z} Q(x) e^{\alpha U(x)},
    \qquad \text{where} \qquad
    Z = \sum_x Q(x) e^{\alpha U(x)}.
\end{equation}
Accordingly, the choice pattern of the decision-maker is predicted by the \emph{equilibrium distribution}~$P$. Crucially, the probability distribution~$P$ extremizes the following functional \citep{Callen1985,Keller1998}:

\begin{definition}[Negative Free Energy Difference]
Let $Q$ be a probability distribution and let $U$ be a real-valued utility function over the set $\set{X}$. For any $\alpha \in \reals$, define the \emph{negative free energy difference} $F_\alpha[P]$ as
\begin{equation}\label{eq:free-energy}
    F_\alpha[P]
    := \sum_x P(x) U(x)
        - \frac{1}{\alpha} \sum_x P(x) \log \frac{P(x)}{Q(x)}.
\end{equation}
The parameter $\alpha$ is called the inverse temperature.
\end{definition}
Although strictly speaking, the functional $F_\alpha[P]$ corresponds to the negative free energy difference, we will  refer to it as the ``free energy'' in the following for simplicity. When inserting the equilibrium distribution~\eqref{eq:equilibrium} into~\eqref{eq:free-energy}, the extremum of $F_\alpha$ yields:
\begin{equation}\label{eq:extremum}
    \frac{1}{\alpha} \log \biggl( \sum_x Q(x) e^{\alpha U(x)} \biggr).
\end{equation}
For different values of $\alpha$, this extremum takes the following limits:
\begin{align*}
    \lim_{\alpha \rightarrow \infty} \tfrac{1}{\alpha} \log Z
        &= \max_x U(x)
        &&\text{(maximum node)}\\
    \lim_{\alpha \rightarrow 0} \tfrac{1}{\alpha} \log Z
        &= \sum_x Q(x) U(x)
        &&\text{(chance node)}\\
    \lim_{\alpha \rightarrow -\infty} \tfrac{1}{\alpha} \log Z
        &= \min_x U(x)
        &&\text{(minimum node)}
\end{align*}
The case $\alpha \rightarrow \infty$ corresponds to the perfectly rational agent, the case $\alpha \rightarrow 0$ corresponds to the expectation at a chance node and the case $\alpha \rightarrow -\infty$ anticipates the perfectly rational opponent. Therefore, the single expression $\frac{1}{\alpha} \log Z$ can represent the maximum, expectation and minimum depending on the value of $\alpha$.

The inspection of~\eqref{eq:free-energy} reveals that the free energy encapsulates a fundamental decision-theoretic trade-off: it corresponds to the expected utility, penalized---or regularized---by the information cost of transforming the base distribution $Q$ into the final distribution $P$. The inverse temperature plays the role of the conversion factor between units of information and units of utility.

If we want to change the temperature $\alpha$ to $\beta$ while keeping the equilibrium and reference distributions equal, then we need to change the corresponding utilities from $U$ to $V$ in a manner given by the following theorem. Temperature changes will be important for the application of the free energy principle to the general decision trees in Section~\ref{sec:general-decision-trees}.

\begin{theorem}\label{theo:temp-change}
Let $P$ be the equilibrium distribution for a given inverse temperature~$\alpha$, utility function~$U$ and reference distribution~$Q$. If the temperature changes to~$\beta$ while keeping~$P$ and~$Q$ fixed, then the utility function changes to
\[
    V(x) = U(x) - \Bigl( \tfrac{1}{\alpha} - \tfrac{1}{\beta} \Bigr)
        \log \frac{P(x)}{Q(x)}.
\]
\end{theorem}

\subsection{Resource Costs}

Consider the problem of picking the largest number in a sequence $U_0, U_1, U_2, \ldots$ of i.i.d. data, where each $U_i \in \set{U}$ is drawn from a source with probability distribution $M$. After $\alpha$ draws the largest number will be given by $\max \{ U_1, U_2, \ldots, U_\alpha \}$.
Naturally, the larger the number of draws, the higher the chances of observing a large number.

\begin{theorem}\label{theo:temperature-interpretation}
Let $\set{X}$ be a finite set. Let $Q$ and $M$ be strictly positive probability distributions over $\set{X}$. Let $\alpha$ be a positive integer. Define $M_\alpha$ as the probability distribution over the maximum of $\alpha$ samples from $M$. Then, there are strictly positive constants $\delta$ and $\xi$ depending only on $M$ such that for all $\alpha$,
\[
    \left|
        \frac{ Q(x) e^{\alpha U(x)} }
        { \sum_{x'} Q(x') e^{\alpha U(x')} }
        - M_\alpha(x)
    \right|
    \leq e^{-(\alpha-\xi) \delta}.
\]
\end{theorem}

Consequently, one can interpret the inverse temperature as a resource parameter that determines how many samples are drawn to estimate the maximum. Note that the distribution~$M$ is arbitrary as long as it has the same support as $Q$. This interpretation can be extended to a negative $\alpha$, by noting that $\alpha U(x) = (-\alpha)(-U(x))$, i.e.\ instead of the maximum we take the minimum of $-\alpha$ samples.

\section{General Decision Trees}\label{sec:general-decision-trees}

A generalized decision tree is a tree where each node corresponds to a possible interaction history $x_{\leq t} \in \set{X}^t$, where $t$ is smaller or equal than some fixed horizon $T$, and where edges connect two consecutive interaction histories. Furthermore, every node $x_{\leq t}$ has an associated inverse temperature $\beta(x_{\leq t})$; and every transition has a base probability $Q(x_t|x_{<t})$ of moving from state $x_{<t}$ to state $x_{\leq t} = x_{<t} x_t$ representing the stochastic law the interactions follow when it is not controlled, and an immediate reward $R(x_t|x_{<t})$. The objective of the agent is to make decisions such that the sum $\sum_{t=1}^T R(x_t|x_{<t})$ is maximized subject to the temperature constraints.

\subsection{Free Energy for General Decision Trees}

The free energy principle is stated above for one decision variable $x$. If $x$ represents a tuple of (possibly dependent) random variables $x_1, \ldots, x_T$, then the free energy principle can be applied in a straightforward manner to the corresponding tree. However, all nodes of the tree will have the same inverse temperature assigned to them and, therefore, the same amount of computational resources will be spent at each node of the tree. This allows for example deriving the formalisms of path integral control and KL control \citep{Todorov2009, BraunOrtega2011, Kappen2012}.

\begin{figure}[htbp]
\floatconts
  {fig:transformation}
  {\caption{The free energy formalism can only be applied in a straightforward manner to trees with uniform resource allocation (left). In order to apply it to general trees that have different resource parameters at each node (right), we need to transform the utilities as described in~\eqref{eq:rewards} to preserve the equilibrium distribution.}}
  {\scriptsize
  \psfrag{a1}[c]{$\alpha$}
  \psfrag{a2}[c]{$\alpha$}
  \psfrag{a3}[c]{$\alpha$}
  \psfrag{b1}[c]{$\beta(\varepsilon)$}
  \psfrag{b2}[c]{$\beta(0)$}
  \psfrag{b3}[c]{$\beta(1)$}
  \psfrag{s1}[c]{$S(0|\varepsilon)$}
  \psfrag{s2}[c]{$S(1|\varepsilon)$}
  \psfrag{s3}[c]{$S(0|0)$}
  \psfrag{s4}[c]{$S(1|0)$}
  \psfrag{s5}[c]{$S(0|1)$}
  \psfrag{s6}[c]{$S(1|1)$}
  \psfrag{r1}[c]{$R(0|\varepsilon)$}
  \psfrag{r2}[c]{$R(1|\varepsilon)$}
  \psfrag{r3}[c]{$R(0|0)$}
  \psfrag{r4}[c]{$R(1|0)$}
  \psfrag{r5}[c]{$R(0|1)$}
  \psfrag{r6}[c]{$R(1|1)$}
  \psfrag{u1}[c]{$U(\varepsilon)$}
  \psfrag{u2}[c]{$U(0)$}
  \psfrag{u3}[c]{$U(1)$}
  \psfrag{v1}[c]{$V(\varepsilon)$}
  \psfrag{v2}[c]{$V(0)$}
  \psfrag{v3}[c]{$V(1)$}
  \includegraphics[]{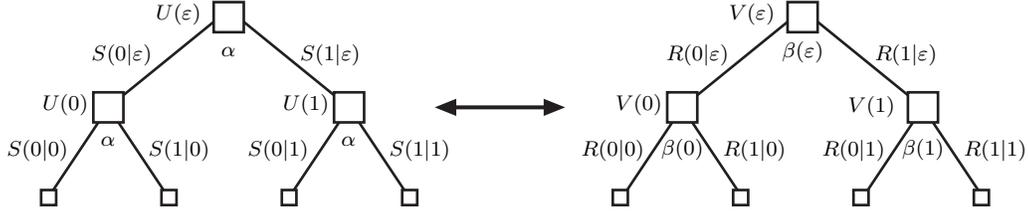}}
\end{figure}

In the case of general decision trees the assumption of uniform temperatures has to be relaxed (Figure~\ref{fig:transformation}). In general, we can then dedicate different amounts of computational resources to each node of the tree. However, this requires a translation between a tree with a single temperature and to a tree with different temperatures. This translation can be achieved using Theorem~\ref{theo:temp-change}. Define a reward as the change in utility of two subsequent nodes. Then, the rewards of the resulting decision tree are given by
\begin{equation}\label{eq:rewards}
    R(x_t|x_{<t}) :=  \bigl[ V(x_{\leq t}) - V(x_{<t}) \bigr]
    = \bigl[ U(x_{\leq t}) - U(x_{<t}) \bigr]
        - \Bigl( \tfrac{1}{\alpha} - \tfrac{1}{\beta(x_{<t})} \Bigr)
            \log\frac{ P(x_t|x_{<t}) }{ Q(x_t|x_{<t})}.
\end{equation}
This allows introducing a collection of node-specific (not necessarily time-specific) inverse temperatures $\beta(x_{<t})$, allowing for a greater degree of flexibility in the representation of information costs. The next theorem states the connection between the free energy and the general decision tree formulation.

\begin{theorem}\label{theo:free-energy-trajectory}
The free energy of the whole trajectory can be rewritten in terms of rewards:
\begin{align}
    \nonumber
    F_\alpha[P] &=
    \sum_{x_{\leq T}} P(x_{\leq T})
    \biggl\{ U(x_{\leq T})
        - \frac{1}{\alpha} \log \frac{P(x_{\leq T})}{Q(x_{\leq T})} \biggr\}
    \\ \label{eq:seq-free-energy}
    &= U(\varepsilon) +
    \sum_{x_{\leq T}} P(x_{\leq T})
        \sum_{t=1}^T \biggl\{
        R(x_t|x_{<t})
        - \frac{1}{\beta(x_{<t})} \log \frac{P(x_t|x_{<t})}{Q(x_t|x_{<t})} \biggr\}.
\end{align}
\end{theorem}

This translation allows applying the free energy principle to each node with a different resource parameter $\beta(x_{<t})$. By writing out the sum in \eqref{eq:seq-free-energy}, one realizes that this free energy has a nested structure where the latest time step forms the innermost variational problem and all other variational problems of the previous time steps can be solved recursively by working backwards in time. This then leads to the following solution:

\begin{theorem}\label{theo:free-energy-solution}
The solution to the free energy in terms of rewards is given by
\[
    P(x_t|x_{<t}) = \frac{1}{Z(x_{<t})}
        Q(x_t|x_{<t}) \exp\Bigl\{
        \beta(x_{<t}) \bigl[ R(x_t|x_{<t})
            + \frac{1}{\beta(x_{\leq t}) } \log Z(x_{\leq t})
            \bigr] \Bigr\},
\]
where $Z(x_{\leq T}) = 1$ and where for all $t<T$
\[
    Z(x_{<t}) = \sum_{x_t} Q(x_t|x_{<t}) \exp\Bigl\{
        \beta(x_{<t}) \bigl[ R(x_t|x_{<t})
            + \frac{1}{\beta(x_{\leq t}) } \log Z(x_{\leq t})
            \bigr] \Bigr\}.
\]
\end{theorem}

\subsection{Generalized Optimality Equations}

Theorem~\ref{theo:free-energy-solution} together with the properties of the free energy extremum~\eqref{eq:extremum} suggest the following definition.
\begin{definition}[Generalized Optimality Equations]
\label{def:generalizedoptimality}
\[
    V(x_{<t})
    = \frac{1}{\beta(x_{<t})} \log\biggl\{
        \sum_{x_t} Q(x_t|x_{<t}) \exp\Bigl\{
        \beta(x_{<t}) \bigl[ R(x_t|x_{<t})
            + V(x_{\leq t})
            \bigr] \Bigr\} \biggr\}.
\]
\end{definition}
By virtue of our previous analysis, this equation tells us how to recursively calculate the \emph{value function} (i.e.\ the utility of each node) given the computational resources allocated in each node.

It is immediately clear that the three kinds of decision trees mentioned in the introduction are special cases of general decision trees. In particular, the three classical operators are obtained as limit cases:
\[
    V(x_{<t}) =
    \begin{cases}
    \begin{aligned}
        \max_{x_t} \{&R(x_t|x_{<t}) + V(x_{\leq t})\}
            && \text{if $\beta(x_{<t}) = \infty$,} \\
        \vphantom{\max_{x_t}} \expect \{&R(x_t|x_{<t}) + V(x_{\leq t}) \}
            && \text{if $\beta(x_{<t}) = 0$,} \\
        \min_{x_t} \{&R(x_t|x_{<t}) + V(x_{\leq t})\}
            && \text{if $\beta(x_{<t}) = - \infty$.}
    \end{aligned}
    \end{cases}
\]
The familiar Bellman optimality equations for stochastic systems are obtained by considering an agent decision node followed by a random decision node:
\begin{align*}
    V(x_{<t})
    &= \max_{x_t}\Bigl\{ R(x_t|x_{<t}) + V(x_{\leq t}) \Bigr\}
    \\&= \max_{x_t}\Bigl\{ R(x_t|x_{<t})
        + \expect\bigl[ R(x_{t+1}|x_{\leq t}) + V(x_{\leq t + 1})\bigr]  \Bigr\}.
\end{align*}

\section{Discussions \& Conclusions}

Bounded rational decision-making schemes based on the free energy generalize classic decision-making
schemes by taking into account information processing costs measured by the Kullback-Leibler divergence \citep{Wolpert2004,Todorov2009,Peters2010,OrtegaBraun2011b,Kappen2012}. Ultimately, these costs are determined by Lagrange multiplier constraints given by the inverse temperature playing the role of a resource parameter. Here we generalize this approach to general decision trees where each node can have a different resource allocation. Consequently, we obtain generalized optimality equations for sequential decision-making that include the well-known Bellman optimality equation
as well as Expectimax-, Minimax- and Expectiminimax-decision rules depending on the limit values
of the resource parameters. The resource parameters themselves are amenable to interesting computational, statistical and economic interpretations. In the first sense they measure the number of samples needed from a distribution before applying the max operator and therefore correspond directly to computational effort. In the second sense they reflect the confidence of the estimate of the maximum and therefore they can also express risk attitudes. Finally, the resource parameters reflect the control an agent has over a random variable. These different ramifications need to be explored further in the future.

%\acks{%
%This study was supported by the Emmy Noether Grant BR 4164/1-1.
%}%

\appendix
\section{Proofs}
\subsection{Proof of Theorem~\ref{theo:temp-change}}
\begin{proof}
\begin{small}
\[
    \sum_x P(x) U(x) - \frac{1}{\alpha} \sum_x P(x) \log \frac{P(x)}{Q(x)}
    = \sum_x P(x) V(x) - \frac{1}{\beta} \sum_x P(x) \log \frac{P(x)}{Q(x)}
\]
Since the equilibrium and reference distributions~$P(x)$ and~$Q(x)$ are constant but arbitrarily chosen, it must be that
\[
    U(x) - \frac{1}{\alpha} \log \frac{P(x)}{Q(x)}
    = V(x) - \frac{1}{\beta} \log \frac{P(x)}{Q(x)}.
\]
Hence,
\[
    V(x) = U(x) - \Bigl( \tfrac{1}{\alpha} - \tfrac{1}{\beta} \Bigr)
        \log \frac{P(x)}{Q(x)}.
\]
\end{small}
\end{proof}

\subsection{Proof of Theorem~\ref{theo:temperature-interpretation}}
\begin{proof}
\begin{small}
Let $x_1, x_2, \ldots, x_N$ be the ordering of $\set{X}$ such that $U(x_1), U(x_2), \ldots, U(x_N)$. It is well known that the distribution over the maximum of $\alpha$ samples is equal to $F_\alpha(x) = F(x)^\alpha$, where $F$ is the cumulative distribution $F(x_n) = \sum_{k \leq n} M(x_k)$. Defining $F(x_0) := 0$, one has $M_\alpha(x_n) = F(x_n)^\alpha - F(x_{n-1})^\alpha$. Hence, the probability can be bounded as $0 \leq M_\alpha(x_n) \leq F(x_n)^\alpha$, or
\begin{equation}\label{eq:max-bound}
     0 \leq M_\alpha(x_n) \leq e^{-\alpha \gamma(x_n)},
\end{equation}
if we use $F(x_n) = e^{-\gamma(x_n)}$ where $\gamma(x_n) \geq 0$. The Boltzmann distribution can be bounded as
\[
    0
    \leq
    \frac{ Q(x_n) e^{\alpha U(x_n)} }{ \sum_k Q(x_k) e^{\alpha U(x_k)} }
    \leq
    \frac{ Q(x_n) e^{\alpha U(x_n)} }{ Q(x_N) e^{\alpha U(x_N)} }.
\]
The upper bound is obtained by dropping all the summands in the expectation but the largest. In exponential form, the bounds are written as
\begin{equation}\label{eq:boltz-bound}
    0
    \leq
    \frac{ Q(x_n) e^{\alpha U(x_n)} }{ \sum_k Q(x_k) e^{\alpha U(x_k)} }
    \leq
    e^{-\alpha \delta(x_n) + c(x_n)},
\end{equation}
where $\delta(x_n) := U(x_N) - U(x_n)$, $c(x_n) := -\log Q(x_N) + \log Q(x_n)$. Note that $\delta(x_n)$ is positive. Subtracting the inequalities~\eqref{eq:max-bound} from~\eqref{eq:boltz-bound} yields
\[
    - e^{-\alpha\gamma(x_n)}
    \leq \frac{ Q(x_n) e^{\alpha U(x_n)} }
         { \sum_k Q(x_k) e^{\alpha U(x_k)} }
    - M_\alpha(x_n)
    \leq e^{-\alpha\delta(x_n) + c(x_n)}.
\]
Choosing $\xi(x_n) = c(x_n) / \delta(x_n) \geq 0$ allows rewriting the upper bound and changing the lower bound to
\[
    - e^{-(\alpha-\xi(x_n)) \gamma(x_n)}
    \leq \frac{ Q(x_n) e^{\alpha U(x_n)} }
         { \sum_k Q(x_k) e^{\alpha U(x_k)} }
    - M_\alpha(x_n)
    \leq e^{-(\alpha-\xi(x_n)) \delta(x_n)}.
\]
Finally, choosing $\xi := \max_n \{ \xi(x_n) \}$ and $\delta = \max\{ \max_n\{\delta(x_n)\}, \max_n\{\gamma(x_n)\} \}$ yields the bounds of the theorem
\[
    - e^{-(\alpha-\xi) \delta}
    \leq \frac{ Q(x_n) e^{\alpha U(x_n)} }
         { \sum_k Q(x_k) e^{\alpha U(x_k)} }
    - M_\alpha(x_n)
    \leq e^{-(\alpha-\xi) \delta}.
\]
\end{small}
\end{proof}

\subsection{Proof of Theorem~\ref{theo:free-energy-trajectory}}
\begin{proof}
\begin{small}
The free energy of the whole trajectory with inverse temperature $\alpha$ is given by
\[
    \sum_{x_{\leq T}} P(x_{\leq T})
    \biggl\{ U(x_{\leq T})
        - \frac{1}{\alpha} \log \frac{P(x_{\leq T})}{Q(x_{\leq T})} \biggr\}.
\]
Using a telescopic sum $\sum_{t=1}^T (a_t - a_{t-1}) = a_T - a_0$ for the utilities yields
\[
    U(\varepsilon) +
    \sum_{x_{\leq T}} P(x_{\leq T})
        \sum_{t=1}^T \biggl\{
        \bigl[ U(x_{\leq t})- U(x_{<t}) \bigr]
        - \frac{1}{\alpha} \log \frac{P(x_t|x_{<t})}{Q(x_t|x_{<t})} \biggr\}.
\]
Using the definition of rewards~\eqref{eq:rewards}, one gets the result
\[
    U(\varepsilon) +
    \sum_{x_{\leq T}} P(x_{\leq T})
        \sum_{t=1}^T \biggl\{
        R(x_t|x_{<t})
        - \frac{1}{\beta(x_{<t})} \log \frac{P(x_t|x_{<t})}{Q(x_t|x_{<t})} \biggr\}.
\]
\end{small}
\end{proof}

\subsection{Proof of Theorem~\ref{theo:free-energy-solution}}
\begin{proof}
\begin{small}
The inner sum of the free energy
\[
    U(\varepsilon) +
    \sum_{x_{\leq T}} P(x_{\leq T})
        \sum_{t=1}^T \biggl\{
        R(x_t|x_{<t})
        - \frac{1}{\beta(x_{<t})} \log \frac{P(x_t|x_{<t})}{Q(x_t|x_{<t})} \biggr\}.
\]
can be expanded as
\begin{align*}
    U(\varepsilon) +
    &\sum_{x_1} P(x_1) \biggl\{
    R(x_1) - \frac{1}{\beta(\varepsilon)}
        \log \frac{P(x_1)}{Q(x_1)} \\
    &+ \sum_{x_2} P(x_2|x_1) \biggl\{
        R(x_2|x_1) - \frac{1}{\beta(x_1)}
            \log \frac{P(x_2|x_1)}{Q(x_2|x_1)} \\
    &+ \cdots \\
    &+ \sum_{x_T} P(x_T|x_{<T}) \biggl\{
        R(x_T|x_{<T}) - \frac{1}{\beta(x_{<T})}
            \log \frac{P(x_T|x_{<T})}{Q(x_T|x_{<T})}
    \biggr\} \cdots
    \biggr\}
    \biggr\}.
\end{align*}
This can be solved by induction, starting with the innermost sums and then recursively solving the outer sums. The innermost sums
\[
    \sum_{x_T} P(x_T|x_{<T}) \biggl\{
        R(x_T|x_{<T}) - \frac{1}{\beta(x_{<T})}
            \log \frac{P(x_T|x_{<T})}{Q(x_T|x_{<T})}
    \biggr\}
\]
are maximized when
\[
    P(x_T|x_{<T}) = \frac{1}{Z(x_{<T})} Q(x_T|x_{<T})
        \exp\Bigl\{
        \beta(x_{<T}) R(x_T|x_{<T}) \Bigr\}.
\]
This can be seen by noting that for probabilities $p_i$ and positive numbers $r_i>0$, the quantity $\sum_i p_i \log (p_i/r_i)$ is minimized by choosing $p_i = \frac{1}{Z} r_i$, where $Z = \sum_i r_i$ is just a normalizing constant. Substituting this solution yields the outer sums
\[
    \sum_{x_t} P(x_t|x_{<t}) \biggl\{
        R(x_t|x_{<t}) - \frac{1}{\beta(x_{<t})}
            \log \frac{P(x_t|x_{<t})}{Q(x_t|x_{<t})}
            + \frac{1}{\beta(x_{\leq t})} \log Z(x_{\leq t})
    \biggr\}
\]
where
\[
    Z(x_{<t}) = \sum_{x_t} Q(x_t|x_{<t}) \exp\Bigl\{
        \beta(x_{<t}) \bigl[ R(x_t|x_{<t})
            + \frac{1}{\beta(x_{\leq t}) } \log Z(x_{\leq t})
            \bigr] \Bigr\}.
\]
These sums are then maximized by choosing
\[
    P(x_t|x_{<t}) = \frac{1}{Z(x_{<t})} Q(x_t|x_{<t})
        \exp\Bigl\{
        \beta(x_{<t}) \bigl[ R(x_t|x_{<t}) +
        \frac{1}{\beta(x_{\leq t})}
        \log Z(x_{\leq t}) \bigr] \Bigr\}.
\]
\end{small}
\end{proof}

\bibliography{bibliography}

\end{document}